\documentclass[10pt,twocolumn,letterpaper]{article}

\usepackage{cvpr}              % To produce the CAMERA-READY version
\usepackage[dvipsnames]{xcolor}

\usepackage{booktabs} % For better looking tables
\usepackage{multirow} % Required for multirows
\usepackage{adjustbox}
\usepackage{pifont}
\usepackage{float}

\definecolor{cvprblue}{rgb}{0.21,0.49,0.74}
\usepackage[pagebackref,breaklinks,colorlinks,citecolor=cvprblue]{hyperref}

\def\paperID{34} % *** Enter the Paper ID here
\def\confName{CVPR}
\def\confYear{2024}

\title{From Forest to Zoo: Great Ape Behavior Recognition with ChimpBehave}

\author{Michael Fuchs, Emilie Genty, Adrian Bangerter, Klaus Zuberbühler, Paul Cotofrei\\
\\
University of Neuchâtel, Switzerland \\
\\
{\tt\small \{michael.fuchs, emilie.genty, adrian.bangerter, klaus.zuberbuehler, paul.cotofrei\}}\tt\small@unine.ch}
\begin{document}
\maketitle
\begin{abstract}
This paper addresses the significant challenge of recognizing behaviors in non-human primates, specifically focusing on chimpanzees. Automated behavior recognition is crucial for both conservation efforts and the advancement of behavioral research. However, it is significantly hindered by the labor-intensive process of manual video annotation. Despite the availability of large-scale animal behavior datasets, the effective application of machine learning models across varied environmental settings poses a critical challenge, primarily due to the variability in data collection contexts and the specificity of annotations.

In this paper, we introduce ChimpBehave, a novel dataset featuring over 2 hours of video (approximately 193,000 video frames) of zoo-housed chimpanzees, meticulously annotated with bounding boxes and behavior labels for action recognition. ChimpBehave uniquely aligns its behavior classes with existing datasets, allowing for the study of domain adaptation and cross-dataset generalization methods between different visual settings. Furthermore, we benchmark our dataset using a state-of-the-art CNN-based action recognition model, providing the first baseline results for both within and cross-dataset settings. The dataset, models, and code can be accessed at:
\href{https://github.com/MitchFuchs/ChimpBehave}{https://github.com/MitchFuchs/ChimpBehave}

\end{abstract}    
\begin{figure}[htbp]
  \centering
  \includegraphics[width=\linewidth]{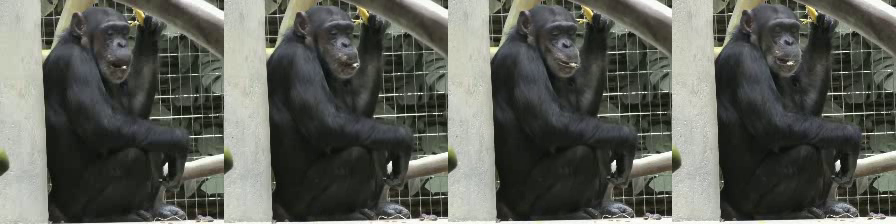}
  \includegraphics[width=\linewidth]{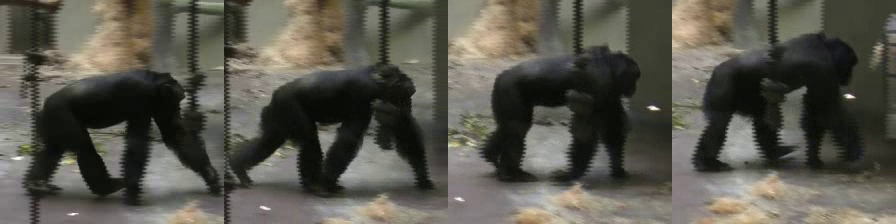}
  \includegraphics[width=\linewidth]{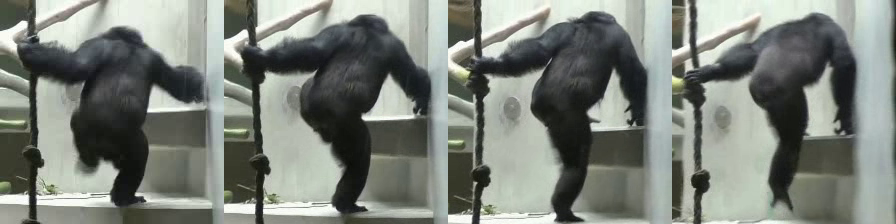}
  \includegraphics[width=\linewidth]{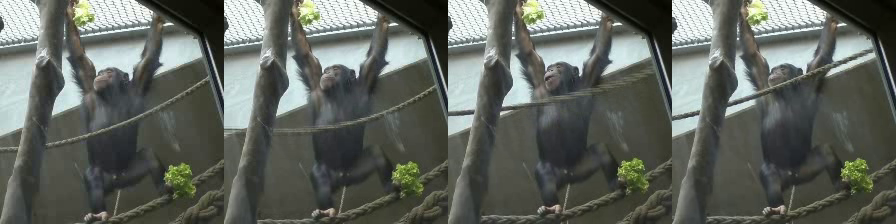}
  \caption{Walking, hanging, sitting, or climbing up? Identifying which chimpanzee behavior is depicted in these images is trivial for most humans. For algorithms, however, this is not always the case, especially when exposed to videos from previously unseen environments. \textbf{TL;DR}: We propose a new dataset and methods to investigate the classification of chimpanzee behaviors across different visual settings.}
  \label{fig:mosaicoleve}
\end{figure}
\section{Introduction}
\label{sec:intro}
The development of machine learning tools to recognize animal behaviors from videos plays a critical role in ecology and ethology. Automated systems for recognizing chimpanzee behaviors could offer a broad spectrum of applications, from enhancing conservation efforts to providing valuable insights into the behavior of great apes. Furthermore, non-invasive technologies developed for their well-being can significantly benefit chimpanzees, an endangered species, in both wild and captive settings. For example, these systems could monitor population dynamics in natural habitats or timely signal behavioral abnormalities in unwell individuals to caretakers in zoos.

As one of humans' closest living relatives, chimpanzees have been the subject of extensive scientific research in fields such as ecology, comparative cognition, neuroscience, and evolutionary biology. This research often relies on videos, whose manual annotation can be time-consuming and labor-intensive. The advancement of algorithms in animal behavior classification can therefore significantly benefit researchers by speeding up the labeling process and/or reducing its overall cost.

Large animal datasets have recently been created to adapt human-centered action recognition models for animal behavior classification (see e.g., \cite{ng2022animal, MammalNet_Chen_2023_CVPR}). 
Although comprehensive, these datasets lack the fine-grained annotations needed to capture the complex behaviors of great apes.
To address this, more focused datasets like ChimpACT \cite{ChimpACT_2023} and PanAf \cite{brookes2024panaf20k} have been created, targeting species-specific behaviors in different environments - from zoo settings to wild forests - and under distinct filming conditions (see \cref{sec:nhp_datasets} for details). 
While valuable, these datasets have two main issues. First, they typically have their own unique set of annotated behaviors. Second, they are often captured under very different conditions (e.g., fixed vs. moving cameras; zoo vs. forest environments). As a result, their joint exploitation for developing action recognition systems that work in diverse conditions is quite challenging.

To address these challenges, we make the following contributions:
\begin{itemize}
\item 
We introduce ChimpBehave, a dataset for great ape behavior recognition, which features over 2 hours of video (approximately 193,000 video frames) annotated with fine-grained behaviors and bounding boxes. Its label classes are specifically aligned with existing datasets to facilitate the study of domain adaptation techniques and cross-dataset generalization experiments.
\item 
We establish a first comparative baseline using X3D \cite{feichtenhofer2020x3d}, a state-of-the-art CNN-based model, in different scenarios in both within-dataset and cross-dataset settings. We use a rigorous evaluation protocol including a stratified 5-fold cross-validation procedure to validate our results.
\end{itemize}

\begin{figure}[htbp]
  \centering
  \includegraphics[width=\linewidth]{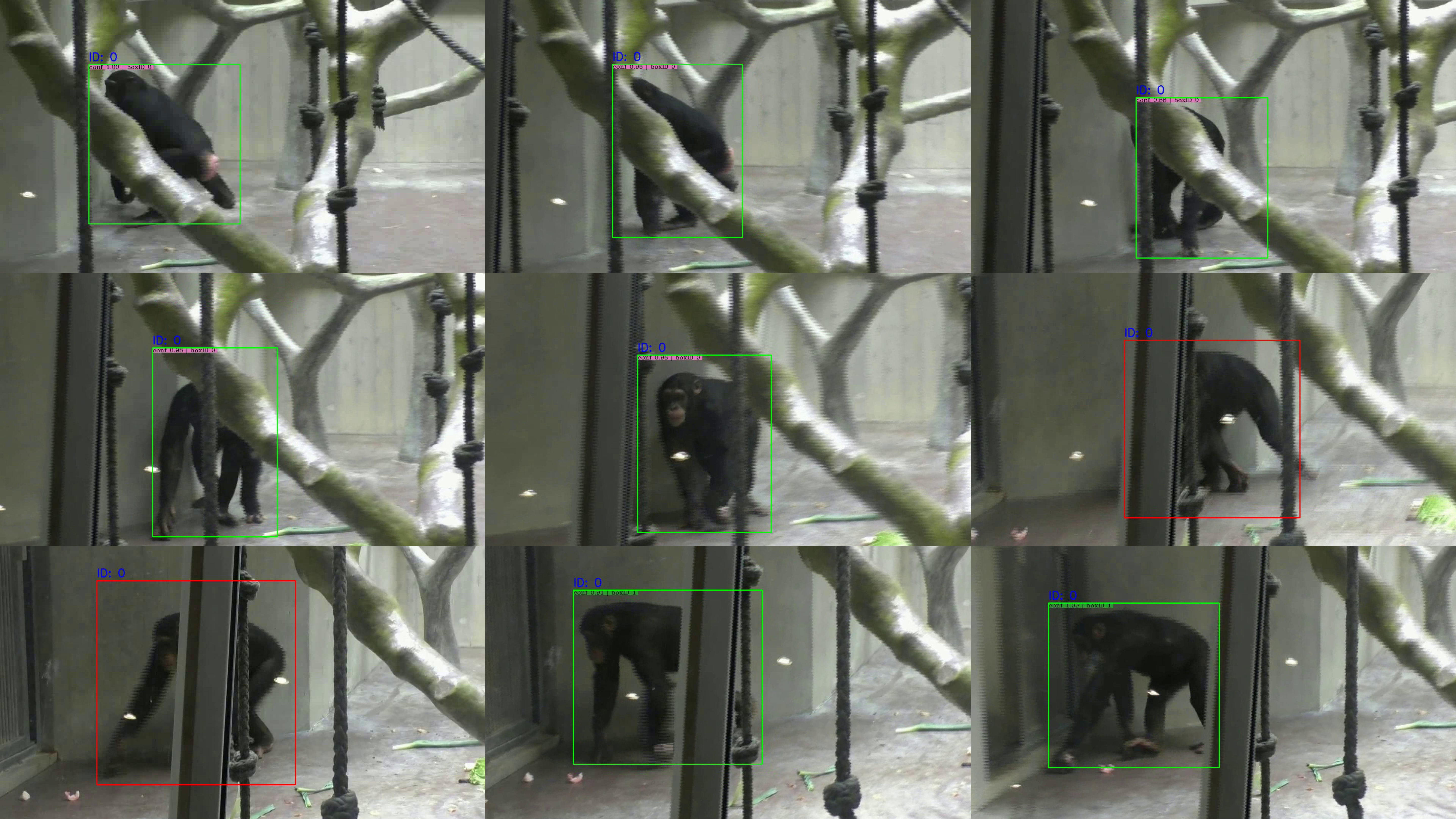}
  \caption{Tracking example after interpolation for missing frames (red bounding box) and corrected ID. As the individual passes behind the pole, its track was lost and the ID was swapped with another.}
  % \caption{Sample images from the ChimpBehave dataset, for six different behaviors, namely 'sitting', 'standing', 'walking', 'running', 'climbind down' and 'climbing up' (in reading order).}
  \label{fig:mosaic_tracking}
\end{figure}

\section{Related work}
\label{sec:relatedwork}
%-------------------------------------------------------------------------

\subsection{Non-human primate datasets}
\label{sec:nhp_datasets}
The growing number of animal datasets designed for computer vision tasks has significantly included non-human primates (NHP), reflecting their significance across various biological and ecological studies. They span a wide array, from encompassing multiple animal orders ~\cite{ng2022animal, MammalNet_Chen_2023_CVPR, yang2023aptv2, Liu_2023_ICCV} to focusing specifically on primates ~\cite{yao2023openmonkeychallenge}, apes ~\cite{ChimpACT_2023, brookes2024panaf20k, desai2022openapepose}, monkeys ~\cite{caged_monkey_2022}, and particularly macaques ~\cite{bala2020automated, labuguen2021macaquepose, macaction_Martini2024}. \\
These datasets showcase notable diversity in annotations and tasks, including  species identification ~\cite{MammalNet_Chen_2023_CVPR, yang2023aptv2, yao2023openmonkeychallenge, brookes2024panaf20k, desai2022openapepose}, animal detection and tracking  ~\cite{yang2023aptv2, ChimpACT_2023, brookes2024panaf20k, Liu_2023_ICCV}, pose estimation ~\cite{ng2022animal, yang2023aptv2, yao2023openmonkeychallenge, ChimpACT_2023, desai2022openapepose, caged_monkey_2022, bala2020automated, labuguen2021macaquepose, macaction_Martini2024, Liu_2023_ICCV}, and behavior recognition ~\cite{ng2022animal, MammalNet_Chen_2023_CVPR, ChimpACT_2023, brookes2024panaf20k, bala2020automated, macaction_Martini2024, Liu_2023_ICCV}. \\
Notably, ChimpACT ~\cite{ChimpACT_2023} and PanAf ~\cite{brookes2024panaf20k} stand out for their focus on great ape behavior, each adopting distinct approaches and highlighting the challenges of applying computer vision across diverse settings. These datasets exhibit significant differences in both their visual environments and the scope of behaviors annotated. \mbox{ChimpACT} documents the life of a young chimpanzee within a zoo environment, characterized by man-made backgrounds, dynamic, hand-held camera work, and a longitudinal focus on a single individual. In contrast, PanAf offers insights into the lives of chimpanzees and gorillas in their natural habitats, featuring static cameras placed in African forests and capturing a wide array of ape populations. These varying contexts highlight the challenges that computer vision applications can encounter across diverse settings. \\
Additionally, their behavior annotations diverge: \mbox{ChimpACT} captures broader locomotive behaviors and social interactions, whereas PanAf specifies actions like 'climbing up', 'running', and 'standing'. This disparity complicates cross-datasets analyses due to mismatches in annotated behaviors, challenging direct model comparison. \\
ChimpBehave aims to bridge these gaps by combining a visual and filming setup reminiscent of ChimpACT with a behavioral annotation scheme aligned with PanAf’s detailed action categories. 

\subsection{Behavior recognition for non-human primates}
Studies on automated behavior recognition in non-human primates have primarily concentrated on macaques ~\cite{marks2022deep, li2023deep, bala2020automated}, monkeys ~\cite{Liu_2023_ICCV} and apes ~\cite{ChimpACT_2023, brookes2024panaf20k, Sakib_2021, bain2021, fuchs2023asbar, brookes2023triple}. \\
Key advancements in this domain leverage action recognition techniques, which are critical for classifying behaviors from video sequences. These techniques are categorized into three main approaches: video-based, skeleton-based and multi-modal. Video-based action recognition analyzes the visual features directly from the raw video data, capturing movements and interactions within the frame's pixel data ~\cite{ChimpACT_2023, Sakib_2021, brookes2024panaf20k, marks2022deep, Liu_2023_ICCV, brookes2023triple}. In contrast, skeleton-based approaches focus on the movement of key body points or joints, constructing a skeletal representation of the subject to discern specific actions or behaviors ~\cite{fuchs2023asbar, bala2020automated}. Additionally, some studies enhance behavior classification by also incorporating multimodal signals, such as audio cues in  ~\cite{bain2021}, providing a more comprehensive analysis by combining visual movement patterns with relevant sounds and/or vocalizations. 
\section{The ChimpBehave dataset}
\label{sec:chimpbehave}

In this section, we present the dataset we have built and its main features.

\noindent\textbf{Dataset Description:} ChimpBehave consists of 1,121 annotated video segments, derived from 50 focal sampling videos recorded in 2016 at the Basel Zoo indoor enclosure (see \cref{fig:mosaicoleve} and other examples in \cref{fig:mosaic_chimpbehave} in Supplementary Materials). Each video follows one of nine chimpanzees to ensure focused observation of the individual, while still capturing its surroundings and other conspecifics. The filming conditions are naturalistic and include camera motion, zooming, and shaking, similar to \cite{ChimpACT_2023}. All recordings were made with a hand-held camera in 1920x1080 pixel resolution at 25 fps.

\noindent\textbf{Behavior Annotation:} An expert primatologist (EG) meticulously labeled the dataset's behavior annotations in all video segments using ELAN (https://archive.mpi.nl/tla/elan), leading to a total of approximately 193,000 annotated video frames. The annotator focused on seven behavioral classes which represent common primate behaviors, namely 'sitting', 'standing', 'walking', 'running', 'hanging', 'climbing down', and 'climbing up'. These classes were intentionally selected to match those annotated in \cite{brookes2024panaf20k}, in order to facilitate cross-dataset analysis.

\noindent\textbf{Bounding Box Annotations:} A single annotator (MF) labeled approximately 12,000 video frames from 131 video segments with bounding boxes at intervals of every tenth frame, with annotations for the remaining frames interpolated. All annotations were made in Label Studio (https://labelstud.io/) on video segments covering a variety of scenes. For each chimpanzee, a minimum of 500 frames were annotated, and for each of the 50 original focal videos, at least two segments or 100 frames were included. These labels were later used to fine-tune QDTrack \cite{pang2021quasi} on the MMaction2 platform \cite{MMAction2_Contributors_OpenMMLab_s_Next_Generation_2020}, a state-of-the-art Multiple Object Tracking (MOT) model pretrained on the ChimpACT dataset \cite{ChimpACT_2023}. This particular model was selected due to its demonstrated adequacy for this task as highlighted in \cite{ChimpACT_2023}. Fine-tuning was conducted for 2 epochs using 107 video segments, leaving out 24 video segments for evaluation purposes. The final model achieves the following scores on common MOT metrics at test time: Recall: 0.7640, Precision: 0.9740, HOTA: 0.6630, mAP: 0.8400.

Using this model, we then predicted tracking bounding boxes for all video segments, manually refined the tracks to fix any ID swaps, and interpolated the predictions where the model's predictions were missing (\cref{fig:mosaic_tracking}). Each track was individually reviewed, and only valid sequences were kept as part of the final dataset. The code for MOT fine-tuning and data conversion between MMaction2 and Label Studio will be made available on our code repository.
\begin{figure}[htbp]
  \centering
  \includegraphics[width=\linewidth]{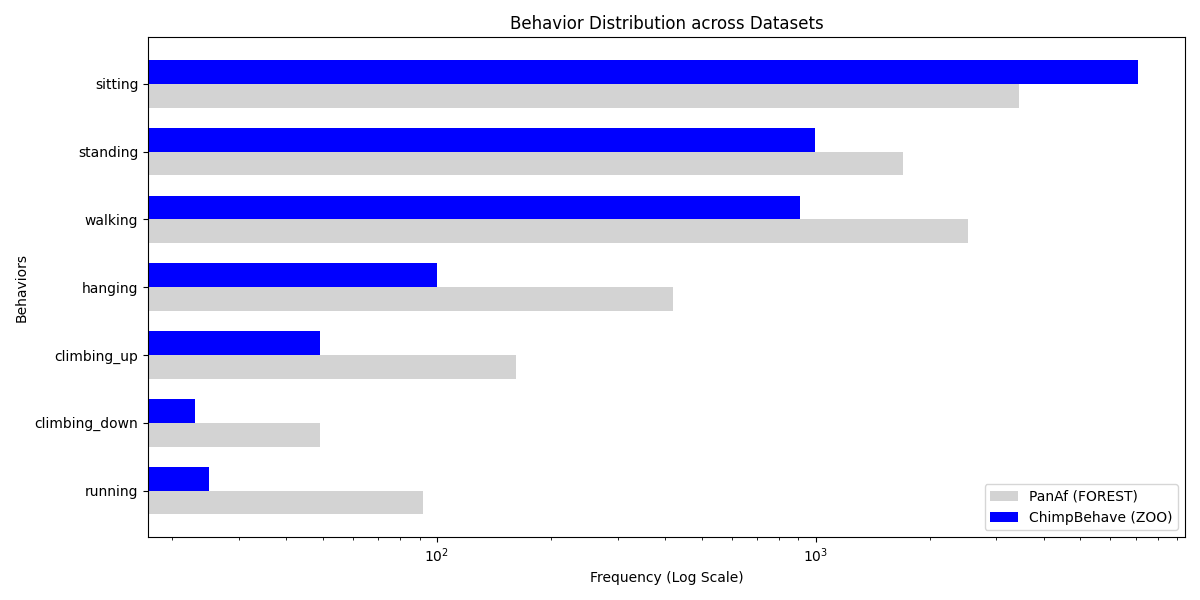}
  \caption{Behavior frequency distribution in the ChimpBehave and PanAf datasets, used in our experiments. Frequencies are plotted on a logarithmic scale to highlight the long-tailed characteristic of the data}
  \label{fig:distribution}
\end{figure}
\noindent\textbf{Data Preprocessing:} For our experiments, we extracted 9,187 unique \textit{miniclips} (a sequence of 20 consecutive frames, without overlap) from all video segments. All remaining frames in segments were discarded. The coordinates of the individual's bounding boxes within the miniclip are used to calculate the global minimum and maximum coordinates of a cropping area, which is adjusted to ensure a minimum size of 224x224 pixels. This process ensures that each action recognition input includes 20 frames of size 224x224, associated with one behavioral class.

%-------------------------------------------------------------------------

\section{Method and experiments}
\label{sec:methods}
%-------------------------------------------------------------------------
Our goal is to study cross-data generalization performance of a standard action recognition model. 
To this end, we first describe the second dataset we have used, before presenting 
the tested behavior recognition model and our experimental protocol. 

\subsection{PanAf dataset}

The Pan African Programme 'The Cultured Chimpanzee' \cite{panaf} aimed to enhance the understanding of evolutionary-ecological factors influencing chimpanzee behavioral diversity. In its efforts, it collected numerous hours of footage from camera traps placed in the forests of Central Africa. From this collection, 500 videos of chimpanzees or gorillas, each 15 seconds long (180,000 frames at 24 FPS, resolution 720x404), were annotated with bounding box coordinates for ape detection and behaviors for action recognition \cite{brookes2024panaf20k, Sakib_2021} (see image examples in Fig~\ref{fig:mosaic_panaf} in Supplementary Materials). The nine labeled behaviors include 'walking', 'standing', 'sitting', 'running', 'hanging', 'climbing up', 'climbing down', 'sitting on back', and 'camera interaction'.

\noindent\textbf{Data Preprocessing:}
We include in our experiments all videos that are fully annotated and extract from them 8,404 miniclips in a similar fashion as detailed in \cref{sec:chimpbehave} from all behavioral classes except 'camera interaction' and 'sitting on back', which are absent in ChimpBehave.

\begin{figure*}[htbp]
  \centering
  \includegraphics[width=\textwidth]{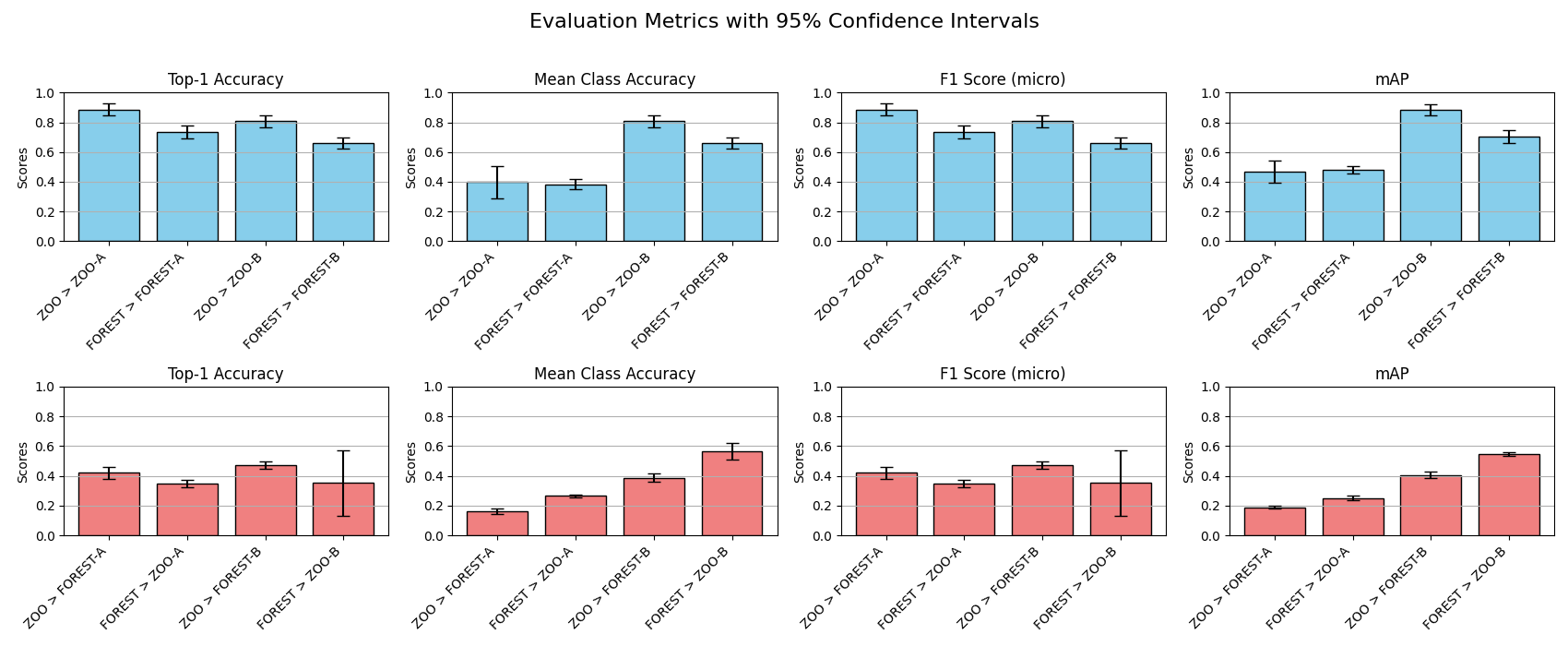}
  \caption{Evaluation metrics on both datasets across both scenarios. The top row (blue) represents the scores within-dataset, while the bottom row (red) represents the scores in cross-dataset settings.}
  \label{fig:evaluation_metrics}
\end{figure*}
\subsection{Behavior Recognition Model}

X3D \cite{feichtenhofer2020x3d}, a state-of-the-art CNN-based model for action recognition, was chosen to conduct our comparative baseline. This model was selected due to its high performance in the PanAf benchmark as presented in \cite{brookes2024panaf20k}. Notably, X3D's architecture expands incrementally from a small 2D image classification model along multiple network axes, including space, time, depth, and width. This network expansion process is designed to find an optimal balance between model complexity and performance, resulting in efficient models without compromising accuracy.     

\subsection{Experimental protocol}

\begin{figure*}[htbp]
  \centering
  \includegraphics[width=\textwidth]{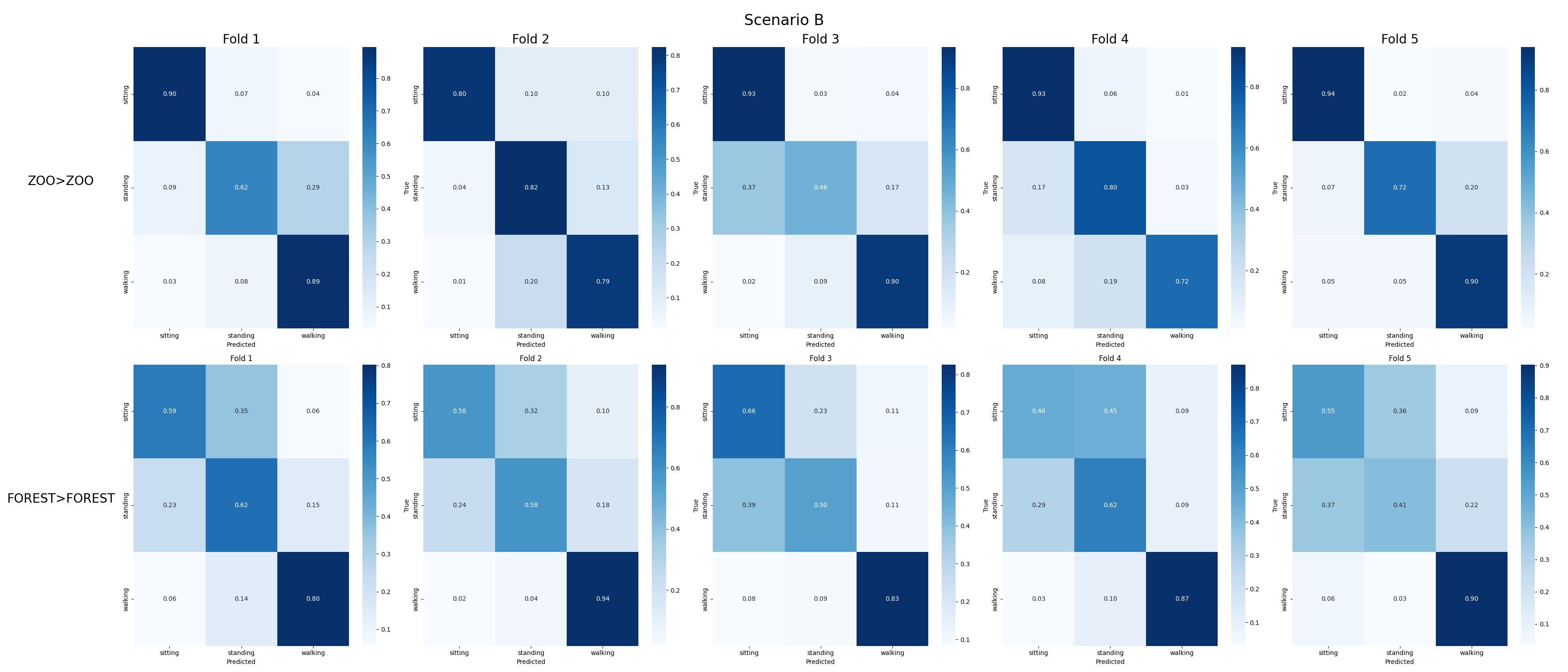}
  \caption{Confusion matrices for each fold in cross-validation: Scenario B - Within-dataset}
  \label{fig:cm_B_within_domain}
\end{figure*}

\begin{figure*}[htbp]
  \centering
  \includegraphics[width=\textwidth]{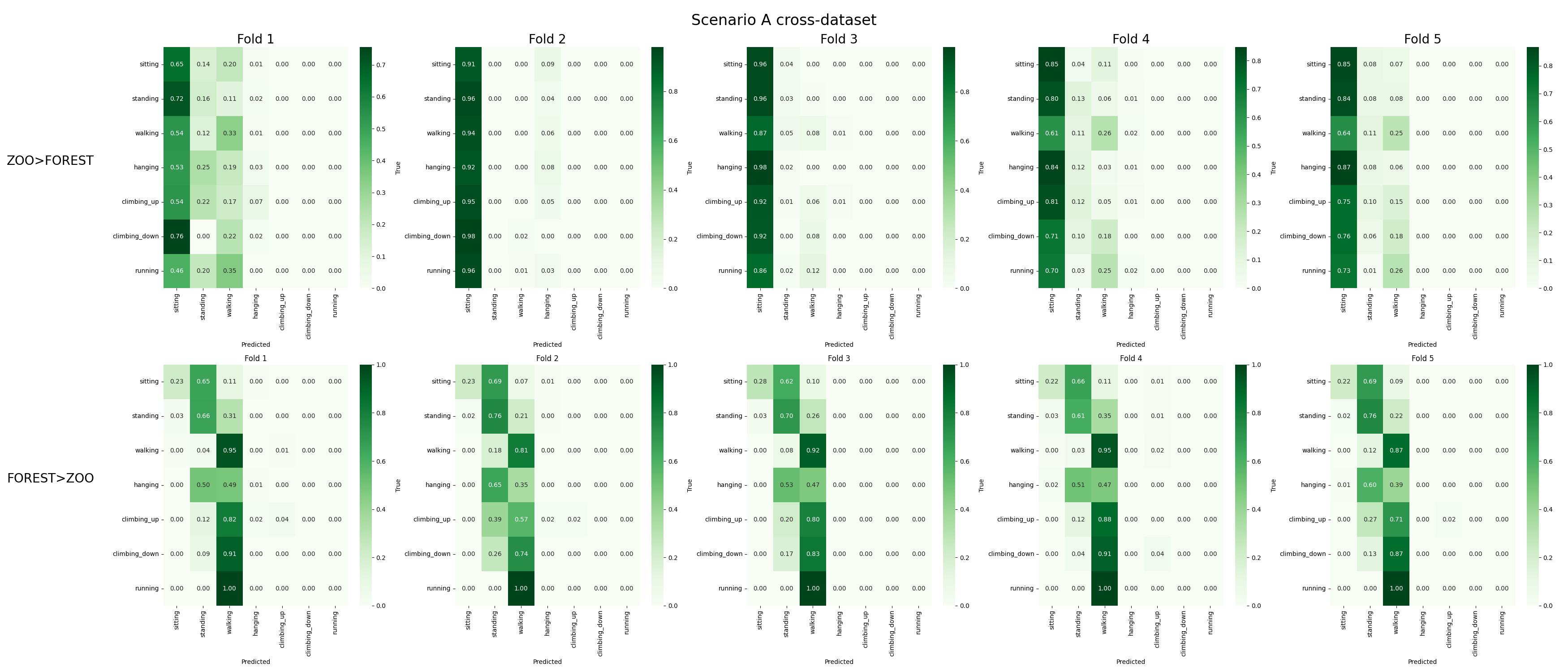}
  \caption{Confusion matrices for each fold in cross-validation: Scenario A - Cross-dataset}
  \label{fig:cm_A_OOD}
\end{figure*}

\begin{figure*}[htbp]
  \centering
  \includegraphics[width=\textwidth]{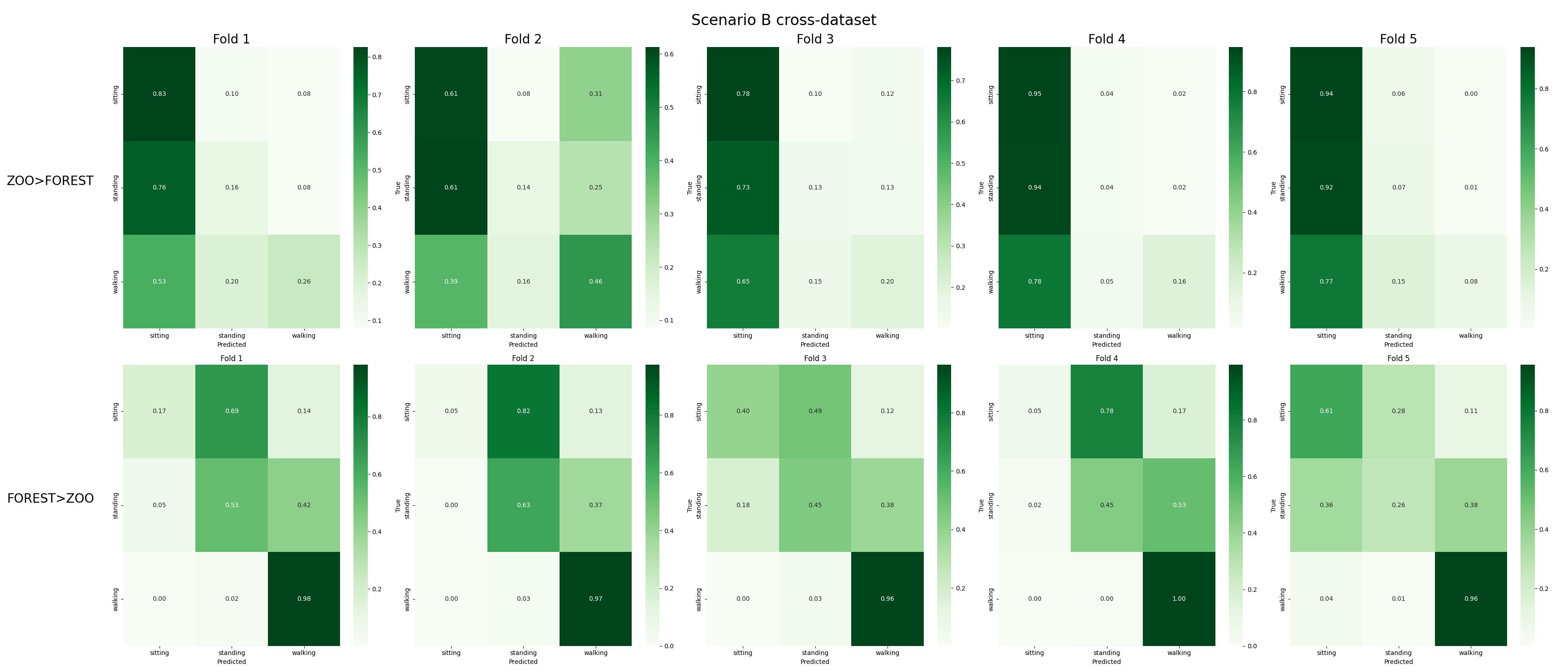}
  \caption{Confusion matrices for each fold in cross-validation: Scenario B - Cross-dataset}
  \label{fig:cm_B_OOD}
\end{figure*}

\noindent\textbf{Evaluation Metrics:} We evaluate each trained model using commonly used action recognition metrics: Top-1 Accuracy, Mean Class Accuracy, F1 (Micro), and Mean Average Precision (mAP). The Top-1 Accuracy metric measures the proportion of correctly classified miniclips, making it sensitive to imbalanced class distributions. F1 (Micro) is particularly useful for imbalanced datasets, as it aggregates the contributions of all classes to compute overall precision, recall, and F1 score, giving equal weight to each miniclip. Mean Class Accuracy evaluates the average accuracy across all classes, ensuring each class is equally represented. Mean Average Precision (mAP) provides a comprehensive measure by averaging the precision across different recall levels for each class, capturing the model's overall ability to identify relevant miniclips. For deeper insights into model predictions, we render all confusion matrices in Supplementary Materials (see \cref{fig:cm_A_within_domain} to \cref{fig:cm_B_OOD}).

\noindent\textbf{Cross-Validation Procedure:}
To validate all experimental results, we follow a standard stratified 5-fold cross-validation procedure. In this approach, each of the five models is trained on 4 folds (representing 80\% of the dataset) and validated on the remaining 20\%. This ensures that all dataset miniclips are used exactly once for validation.
To ensure the exact same class distribution across folds, we compute the largest multiple of 5 in the total number of miniclips for each class and discard the rest. 
The miniclips for each fold were selected in the order of their appearance in the database, which was sorted by video name and then by frame numbering, to group them as much as possible at the video level. This step was taken to ensure maximum generalization across videos, similar to the train/validation/test partitioning followed in \cite{brookes2024panaf20k}.

\noindent\textbf{Confidence Intervals:} 
To assess statistical significance, we calculated 95\% confidence intervals for each metric using the five evaluations from the 5-fold cross-validation procedure, applying a t-distribution ($\alpha=0.025, \nu=4$).

\noindent\textbf{Cross-Dataset Evaluation:} When referring to 'cross-dataset' in our evaluations, we imply that the model was trained on 80\% of the miniclips from one of the datasets (as described above) and later tested on all miniclips of the second dataset.

\noindent\textbf{Comparative Scenarios:} 
As both datasets suffer from class distribution imbalance, we investigate two distinct scenarios referred to as \textbf{A} for \textbf{ALL} and \textbf{B} for \textbf{BALANCED}: 
\begin{itemize}
    \item \textbf{Scenario A} includes all miniclips from all seven classes. 
    \item \textbf{Scenario B} only includes miniclips from the most frequent classes, namely 'sitting', 'standing', and 'walking'. Additionally, in Scenario B, the number of miniclips kept for experiments in each class is set to the same value  across classes and is determined by the number of miniclips in the least frequent class of the three. For the two other classes, a set of miniclips is randomly selected from all its miniclips.
\end{itemize}

\noindent\textbf{Experimental details.}
 Each model training and evaluation was conducted on the MMaction2 platform \cite{MMAction2_Contributors_OpenMMLab_s_Next_Generation_2020} using default hyperparameters. Each X3D model was trained on the HPC cluster of the University of Neuchâtel, on 4x NVIDIA RTX 2080 Ti (4x 11GB) for 50 epochs, using \textit{SGD} optimization, with an initial learning rate of $0.01$. The final epoch was selected based on its Top1 accuracy on the validation set.

\section{Results and discussion}
\label{sec:results}

To emphasize the uniqueness of the visual features of our two datasets and to simplify discussions, in this Section we will refer to the ChimpBehave dataset as ZOO and the PanAf dataset as FOREST.

\noindent\textbf{Within dataset results.}
Our results are shown in \cref{fig:evaluation_metrics}.

When comparing scores between ZOO and FOREST, the model performs statistically significantly better in both scenarios for ZOO with respect to Top-1 Accuracy and F1 Score (micro). Both of these metrics set equal importance on all miniclips. A different pattern emerges when examining mean class accuracy and mAP, two metrics whose final scores are averaged by the number of classes. For these two metrics, we observe no significant difference between ZOO and FOREST in scenario A. This can be partially explained by the long-tailed characteristic of both datasets, where the model systematically fails to classify classes such as 'climbing up,' 'climbing down,' and 'running' across all folds, as seen in \cref{fig:cm_A_within_domain} in Supplementary Materials. However, the model performs significantly better in ZOO across all evaluation metrics, indicating that it learns to recognize visual features in ZOO more easily.

We can identify three major factors that may contribute to this difference. First, when visually comparing sequences of images from both datasets in \cref{fig:mosaic_composite} in Supplementary Materials, one can observe higher contrast between the individual and its background in ZOO. Second, the overall image resolution is markedly higher in ZOO (1920x1080) compared to FOREST (720x404). Third, the relative size of individuals can be much smaller in FOREST, as the camera is fixed, whereas focal sampling videos in ZOO follow each individual.

\noindent\textbf{Evaluating generalization: cross-dataset results.}
The model's generalization capabilities show contrasting patterns, depending on the evaluation metric. For mean class accuracy and mean average precision, the ZOO$\rightarrow$FOREST models obtained statistically significant lower performances than the FOREST$\rightarrow$ZOO models in both scenarios. Conversely, for Top-1 Accuracy and F1-score, the ZOO$\rightarrow$FOREST models achieved statistically significant better performances than the FOREST$\rightarrow$ZOO models, but only in scenario A. Additionally, the performances of all models in the cross-dataset context (independent of scenario or evaluation metric) are approximately 50\% of the same performances in the within-dataset context.
\section{Limitations and future work}
\label{sec:future_work}
As highlighted in the previous sections, ChimpBehave suffers from class distribution imbalance, which we aim to address in future work. In this regard, spatio-temporal action detection models, such as those proposed in \cite{ChimpACT_2023}, could facilitate future data annotation efforts.
On another note, while pose estimation is currently a much-researched topic in animal behavior recognition, especially for NHP, our current results only provide a baseline for video-based approaches. In the future, our work could be extended to include a skeleton-based action recognition baseline using methods similar to the one proposed in \cite{fuchs2023asbar}.
\section{Conclusion}
\label{sec:conclusion}
In this paper, we introduced ChimpBehave, a new dataset which we hope will be useful for future research in automated behavior classification of non-human primates. By specifically aligning the labeling scheme with existing datasets, we were able to demonstrate how to perform cross-dataset evaluations and provide an initial baseline for future research. We hope this will prove helpful for the conservation, study, understanding, and well-being of chimpanzees and great apes in general.
\section{Acknowledgement}
\label{sec:acknowledgement}
We extend our gratitude to the Basel Zoo, its staff, and its director, Adrian Baumeyer, for granting us the opportunity to conduct our data collection within their facilities.\\
Furthermore, we wish to express our appreciation to NCCR Evolving Language, Swiss National Science Foundation Agreement \#51NF40\_180888 and Grant No. CR31I3\_166331 awarded to A.B. and K.Z. for their financial support in data collection and annotation. \\
Special thanks are owed to the members of the SIG \textit{Ape Gestures}, including Daphné Bavelier, Richard Hahnloser, Nianlong Gu, and Remo Nitschke, for their contributions.
Thank you to Jean-Marc Odobez, at Idiap Research Institute, for his dedicated help.  
\section{Ethical Statement}
\label{sec:ethical_statement}
We received ethical agreement for this study from the Commission d'Ethique de la Recherche of the University of Neuchâtel (agreement number: 01-FS-2017) and the Kantonales Veterinäramt BS at Basel Zoo.
% \input{sec/0_abstract}    
% \input{sec/zzz_1_intro}
% \input{sec/zzz_2_formatting}
% \input{sec/zzz_3_finalcopy}
% \newpage
{
    \small
    \bibliographystyle{ieeenat_fullname}
    \bibliography{main}
}

\clearpage
\setcounter{page}{1}
\maketitlesupplementary

% \section{Images}
% \label{sec:rationale}
% 
% \begin{figure*}[H]
%   \centering
%   \includegraphics[scale=0.02]{figs/mosaic_chimpbehave.png}
%   \caption{Example images of chimpBehave}
%   \label{fig:mosaic_chimpbehave}
% \end{figure*}

\begin{figure*}[!htbp]
  \centering
  \includegraphics[width=\textwidth]{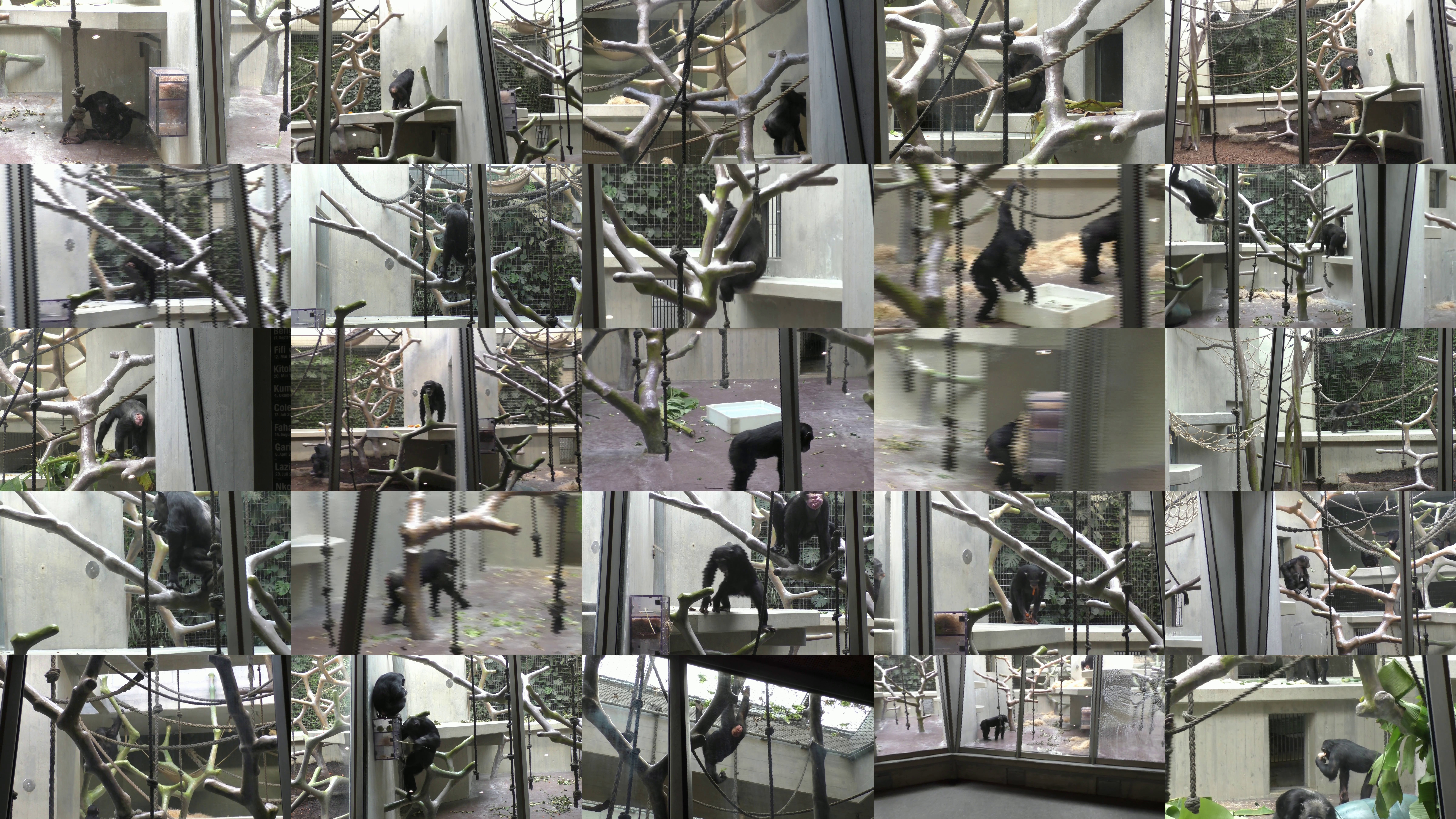}
  \caption{Image examples of ChimpBehave}
  \label{fig:mosaic_chimpbehave}
\end{figure*}

\begin{figure*}[htbp]
  \centering
  \includegraphics[width=\textwidth]{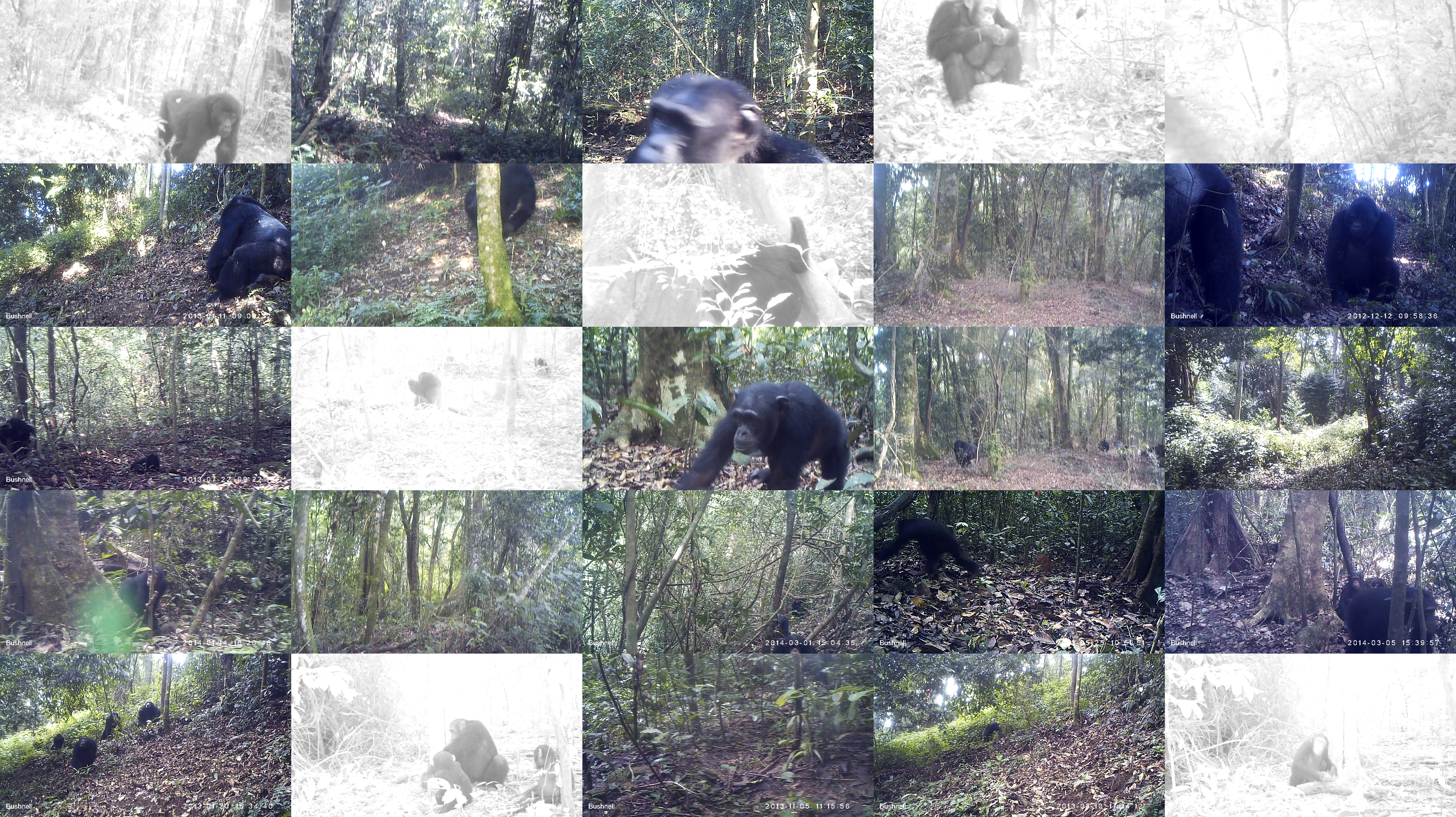}
  \caption{Image examples of PanAf}
  \label{fig:mosaic_panaf}
\end{figure*}

\begin{figure*}[htbp]
  \centering
  \includegraphics[scale=0.2]{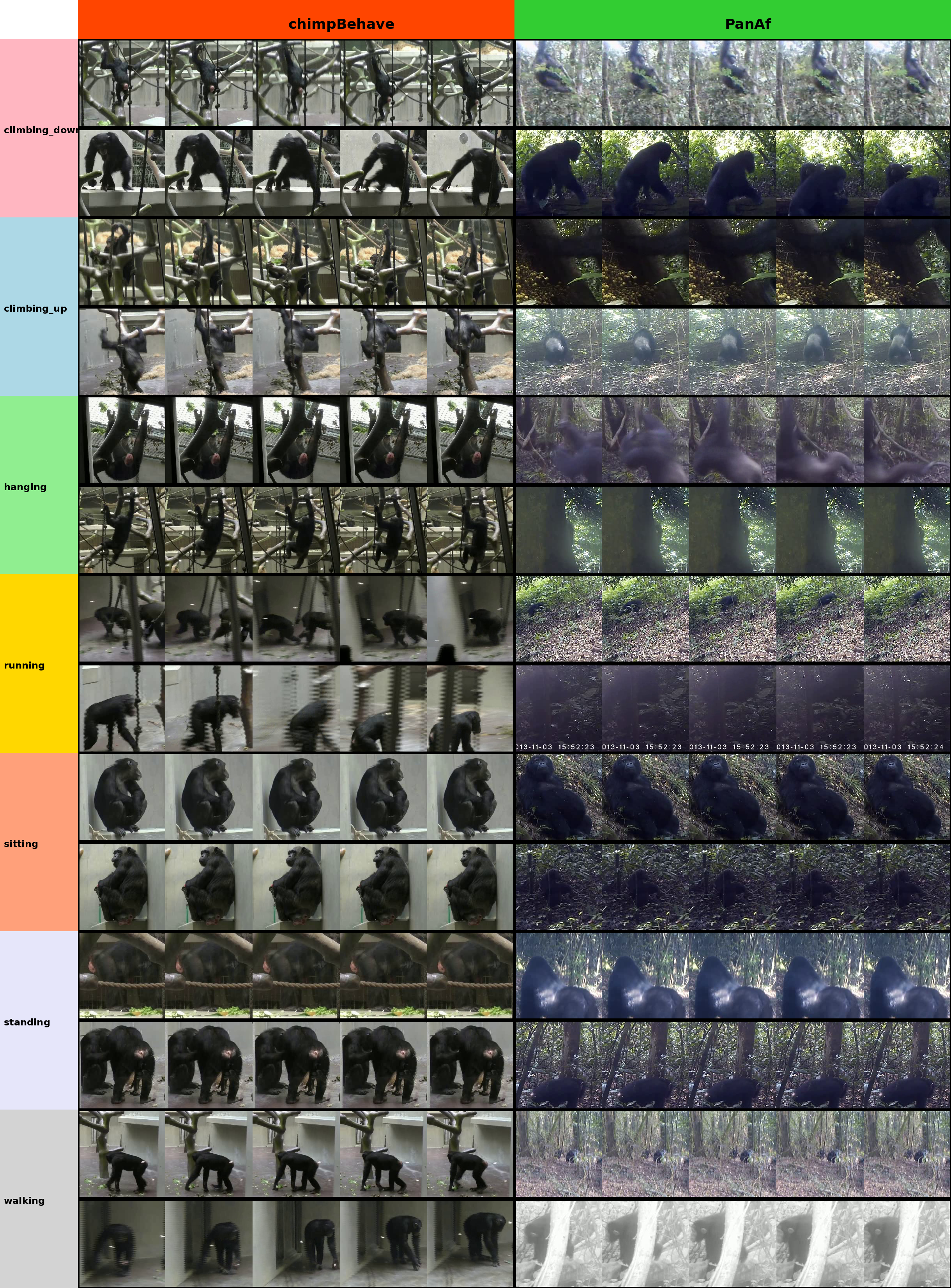}
  \caption{Examples of miniclips between datasets and behavior classes. Note that we sampled 4 out of 20 frames from each miniclip for visualization purposes.}
  \label{fig:mosaic_composite}
\end{figure*}

% \begin{figure*}[htbp]
%   \centering
%   \includegraphics[width=10cm]{figs/within-domain.jpg}
%   \caption{evaluation metrics - within-domain}
%   \label{fig:within_domain_metrics}
% \end{figure*}

% \begin{figure*}[htbp]
%   \centering
%   \includegraphics[width=10cm]{figs/out-of-domain.jpg}
%   \caption{evaluation metrics - out-of-domain}
%   \label{fig:out_of_domain_metrics}
% \end{figure*}

\begin{figure*}[htbp]
  \centering
  \includegraphics[width=\textwidth]{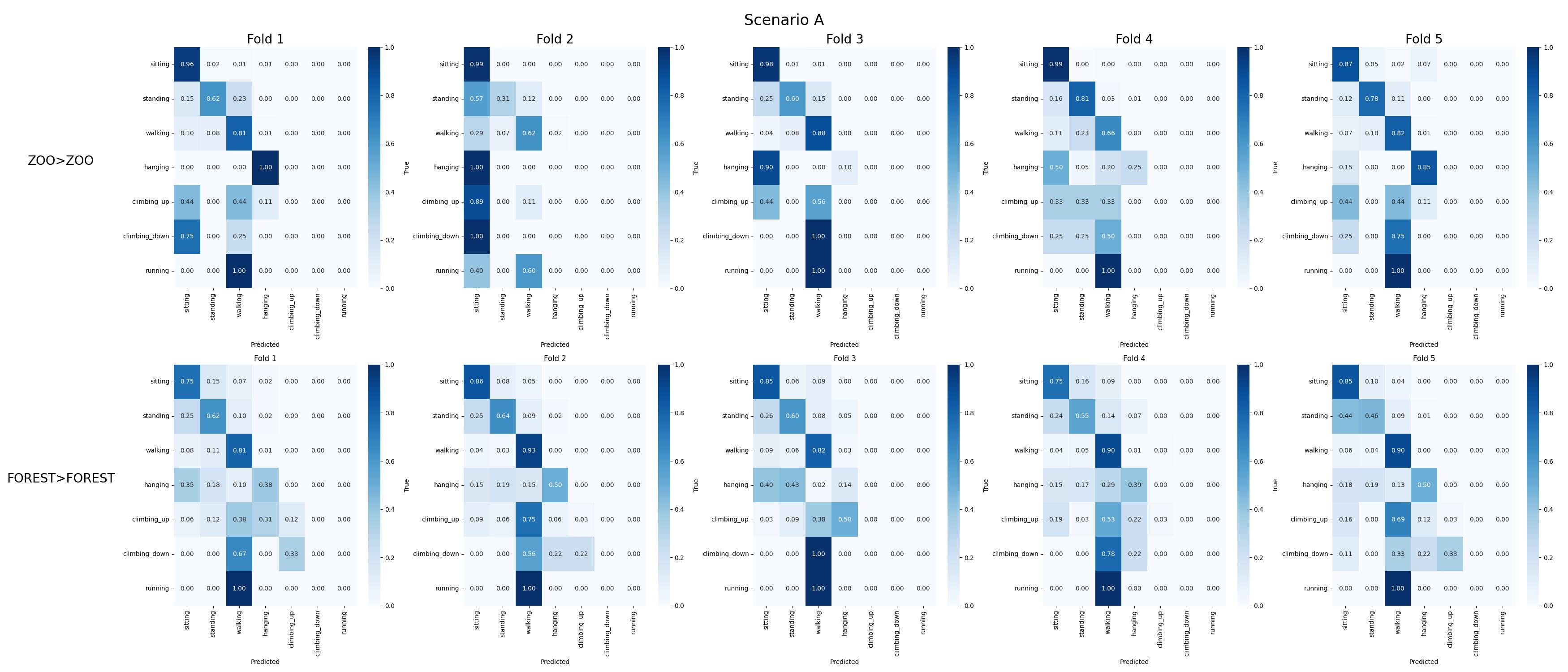}
  \caption{Confusion matrices for each fold in cross-validation: Scenario A - Within-dataset}
  \label{fig:cm_A_within_domain}
\end{figure*}

\end{document}